\documentclass[a4paper]{article}
\usepackage[numbers]{natbib}
\usepackage{helvet}
\usepackage{pdfpages}
\usepackage{graphicx}% http://ctan.org/pkg/graphicx
\usepackage[footnotesize,hang]{caption} % réduire la taille des légendes des images
\usepackage{hyperref}
\usepackage{import}
\usepackage{svg}
\usepackage{tikz}
\usetikzlibrary{positioning}
\usetikzlibrary{calc}
\usepackage{xcolor}
\usepackage{listings}
\usepackage{amsmath}
\usepackage{amssymb}
\usepackage{pifont}
\usepackage{array}
\usepackage{xspace}
\newcolumntype{L}[1]{>{\raggedright\let\newline\\\arraybackslash\hspace{0pt}}m{#1}}
\newcolumntype{C}[1]{>{\centering\let\newline\\\arraybackslash\hspace{0pt}}m{#1}}
\newcolumntype{R}[1]{>{\raggedleft\let\newline\\\arraybackslash\hspace{0pt}}m{#1}}
\date{}
\newcommand{\R}{\mathbb{R}}
\newcommand{\norm}[2]{\left\lVert#1\right\rVert_{#2}}

\newcommand{\smtlib}{SMT-LIB}
\newcommand{\ie}{\emph{i.e.,\xspace}}

\newcommand{\eg}{\emph{e.g.,}\xspace}

\newcommand{\gen}{$g$\xspace}%generator
\newcommand{\per}{$p$\xspace}%perception unit
\newcommand{\rsn}{$r$\xspace}%reasoning unit 
\newcommand{\prm}{$\mathcal{S}$\xspace}%parameter space
\newcommand{\persp}{$\mathcal{X}$\xspace}%perceptive space
\newcommand{\decsp}{$\mathcal{Y}$\xspace}%decision space

\newcommand{\OUTIL}{ONNX2SMT}
\newcommand{\outil}{ONNX2SMT}
\newcommand{\framework}{CAMUS}

\title{CAMUS: A Framework to Build Formal Specifications 
for Deep Perception Systems Using Simulators}

\author{Julien Girard-Satabin\\
    \texttt{julien.girard2@cea.fr}
      \and 
      Guillaume Charpiat\\
      \texttt{guillaume.charpiat@inria.fr}
      \and
      Zakaria Hichem Chihani\\
      \texttt{zakaria.chihani@cea.fr} 
      \and
      Marc Schoenauer\\
      \texttt{marc.schoenauer@inria.fr}
  }
\begin{document}

\maketitle

% \vspace{3cm}
% % NB: color code for comments:
% % \zak{Zak}, \gyo{Guillaume}, \marc{Marc}.
% % \zak{testing}
% % Abstract
\begin{abstract}

The topic of provable deep neural network robustness has raised considerable 
interest in recent years. Most research has focused on adversarial robustness,
which studies the robustness of perceptive models in the neighbourhood of 
particular samples.
However, other works have proved global properties of smaller neural networks. 
Yet, formally verifying perception remains uncharted. This is due notably to 
the lack of relevant properties to verify, as the distribution of possible 
inputs cannot be formally specified.
We propose to take advantage of the simulators often used either to train machine 
learning models or to check them with statistical tests, a growing trend in industry.
Our formulation allows us to formally express and verify safety properties on 
perception units, covering all cases that could ever be generated by the simulator, 
to the difference of statistical tests which cover only seen examples. 
Along with this theoretical formulation, we provide a tool to translate deep 
learning models into standard
logical formulae. As a proof of concept, we train a toy example mimicking 
an autonomous car perceptive unit, and we formally verify that it will never fail to capture the relevant information in the provided 
inputs.
\end{abstract}

% Introduction and related work
\section{Introduction}

Recent years have shown a considerable interest 
in designing ``more robust'' deep learning 
models. In classical software safety, asserting the robustness of a program 
usually consists in checking if the program respects a given specification. 
Various techniques can output a sound answer whether the specification is 
respected or not, provided it is sufficiently formally formulated.
However, the deep learning field is different, since the subject
of verification (the deep learning model) is actually obtained through a 
learning algorithm, which is not tailored to satisfy a specification 
by construction.
In this paper, we will denote as ''program'' the deep learning model that is
the result of a learning procedure. Such a program aims to perform a certain 
task, such as image classification, using statistical inference.
%The learning procedure aims to train a program through a learning algorithm, 
%and usually involve a loss minimization.
For this, it is trained through a learning algorithm, usually involving a loss 
minimization by gradient descent over its parameters.
This way, in most deep learning applications, there exists no formal 
specification of what the program should achieve at the end of the learning 
phase: instead, the current dominant paradigm in statistical learning 
consists in learning an estimator that approximates a probability distribution,
about which little is known. Failures of learning procedures, such as 
overfitting, are hard to quantify and describe in the form of a specification.

A particular flaw of deep learning, namely adversarial examples,
has been the subject of intensive research \cite{szegedy_intriguing_2013,carlini_towards_2016,papernot_transferability_2016,ilyas_adversarial_2019}.
Recently, the quest for provable adversarial robustness
has been bringing together the machine learning
and formal methods communities. New tools are
written, inspired by decades of work in software safety,
opening new perspectives on formal verification for 
deep learning.
However, the bulk of these works has focused on the specific issue of 
adversarial robustness. Apart from well-defined environments where 
strong prior information exists on the input space (see 
\cite{katz_reluplex:_2017}),
little work has been made on formulating and certifying 
specific properties of deep neural networks. 

The goal of this work is 
to propose a framework for
the general problem of deep learning verification
that will allow the formulation of new properties to be checked, while 
still benefiting from the efforts of the formal methods community towards 
more efficient verification tools.
We aim to leverage the techniques developed for adversarial robustness and 
extend the scope of deep learning verification to working on global properties.
Specifically, we focus on a still unexplored avenue: models
trained on simulated data, commonplace in the automotive industry. 
Our contribution is twofold: we first propose a formalism to express 
formal properties on deep perception units trained on simulated data;
secondly we present an open source tool that directly translates 
machine learning models into a logical formula that can be used to soundly 
verify these properties, hence ensuring some formal guarantees. 
Recent work proposed to analyse programs trained on simulators
\cite{dreossi2019verifai}. Although their motivations are similar to ours, 
they work on abstract feature spaces without directly considering the 
perception unit, and they rely on sampling techniques while we aim to use 
sound, exhaustive techniques. Their aim is to exhibit faulty behaviour in 
some type of neural network controllers, while we can formally verify any 
type of perception unit.

The paper is structured as follows: We first describe our formulation
of the problem of verification of machine learning models trained on simulated
data. We then describe the translator tools, and detail its main features. 
Finally, we present as a first use case a synthetic toy 
'autonomous vehicle' problem. We  
conclude by presenting the next issues to tackle.
\section{Related work}
\subsection{Adversarial robustness: a local property}
Adversarial perturbations are small variations of a given example that have been 
crafted so that the network misclassifies the resulting noisy 
example, called an {\em adversarial example}. 
More formally, given a sample $x_0$ %$\mathbf{x_0}$ %% vu qu'aucune autre variable du papier n'est en gras
in a set $\mathcal{X}$,
a classification function $C: %\mathbf{x}
x \in \mathcal{X} \rightarrow \R^d$, 
%a perturbation $\delta_s$, 
a distortion amplitude $\varepsilon \geqslant 0$
and a distance metric~$\norm{.}{p}$, 
a neural network is \textit{locally $\varepsilon$-robust}
  %$\delta_s$-robust}
if for all perturbations $\delta$ s.t.~$\norm{\delta}{p}~\leqslant~\varepsilon, \;\; C(x_0) = C(x_0+\delta)$.
%C(\mathbf{x_0}) = C(\mathbf{x_0+\delta})$.
%%%\norm{\delta_s}{p}$,
To provably assert adversarial robustness of a network, the goal is then
to find the \textit{exact} minimal distortion $\varepsilon$. %$\delta_s$.
Note that this 
property is local, tied to sample $x_0$. %\mathbf{x_0}$.
A global adversarial robustness property could be phrased as:
A deep neural network is \textit{globally} $\varepsilon$-robust if 
for any pair of samples
%$(\mathbf{x_1}, \mathbf{x_2}) \in \mathcal{X}^2$ s.t. $\norm{\mathbf{x_1} - \mathbf{x_2}}{p} \leqslant \varepsilon$, $C(\mathbf{x_1}) = C(\mathbf{x_2})$.
$(x_1, x_2) \in \mathcal{X}^2$ s.t. $\norm{x_1 - x_2}{p} \leqslant \varepsilon$, $C(x_1) = C(x_2)$.
Verifying this global property is intractable,
thus all the work has focused on local adversarial robustness.

Since their initial discovery in \cite{szegedy_intriguing_2013}, adversarial examples
became a widely researched topic.
New ways to generate 
adversarial examples were proposed \cite{carlini_towards_2016,kurakin_adversarial_2016,yao_trust_2018}, as well as defenses
\cite{araujo_robust_2019,madry_towards_2017}.
Other works focus on studying the 
theory behind adversarial examples. While the
initial work \cite{goodfellow_explaining_2014} 
suggests that adversarial examples are a result
of a default in the training procedure, ``bugs'',
recent investigations 
(\cite{ford_adversarial_2019,ilyas_adversarial_2019,simon-gabriel_first-order_2019}) suggest that (at 
least part of the) adversarial examples may be 
inherently linked to the design principles of deep learning and to their 
resulting effects on programs: using any input features available 
to decrease the loss function, 
including ``non robust'' features that are 
exploited by adversarial examples generation 
algorithms. It is important to note that their very existence may be tied with 
the fact that we employ deep neural network on highly-dimensional perceptual
spaces such as images, where we witness counter-intuitive behaviours.
In any case, their imperceptibility for humans 
and their 
capacity to transfer between
networks and datasets 
\cite{papernot_transferability_2016} make 
them a potentially dangerous phenomenon regarding
safety and security. For example, an
autonomous car sensor unit could be fooled by 
a malicious agent to output false direction
in order to cause accidents. 
\subsection{Proving global properties in non-perceptual space}
\label{reluplex}
It is possible to express formal properties in simpler settings than 
adversarial robustness. 
By simpler, we mean two main differences: (i) the 
dimensionality of the input is much lower than in typical perception cases,
where most of adversarial examples occur, and (ii) the problem the program aims 
to solve provides an explicit description of the meaning of the inputs and 
outputs, making a formulation of safety property much simpler.
Rephrased otherwise, the program is working on inputs whose semantics is
(at least partially) defined. 
Since deep neural networks use simple programming concepts (\eg{} no
loops), it is quite easy to translate them directly to a standard verification
format, such as \smtlib{} \cite{barrett_smt-lib_nodate}. 
Provided the inputs are 
sufficiently well defined, it is then possible to encode safety
properties as relationships between inputs and outputs, such as inequality constraints
on real values. 

% For instance, an autonomous car controller receives signals
% with defined meaning (accelerometry, vision) as inputs, and output defined
% commands to actuators. 
An example of such setting can be seen in the Anti Collision Avoidance System
for Unmanned aircrafts (ACAS-Xu) \cite{manfredi2016introduction}.
Inputs correspond to aircraft sensors, and
outputs to airplane commands. In such case, specifications can be directly
encoded as a set of constraints on the inputs and outputs.
In \cite{katz_reluplex:_2017}, the authors proposed an implementation of 
ACAS-Xu as a deep neural network, and they were able to formally prove that 
their program respected various safety properties.

It is important to note that the inputs of the 
program are here high-level information (existence of an intruder together 
with its position), which completely bypasses the problem of perception 
(as airplanes have direct access to this information, in a low-level form, through their sensors, and through %the information provided by the
communications with
ground operators).

\subsection{Tools for provable deep learning robustness}
Critical systems perform operations whose failure may cause physical 
harm or great economical
loss. A self-driving car is a critical system: failure of embedded software
may cause harm, as seen in accidents such as \cite{hawkins_2019}.
In the automobile industry, one expects the airbag to resist to a given
pressure, the tires to last for a lower-bounded duration, etc. 
As software is more and more ubiquitous in vehicles, it is natural to have
high expectations for software safety as well.
An attempt to meet these expectations makes use of formal methods. 
This general term
describes a variety of techniques that aim to provide mathematically sound 
guarantees with respect to a given specification. 
In less than a decade, an impressive 
amount of research was undertaken to
bring formal verification knowledge and tools 
to the field of adversarial robustness.
Deep learning verification has developed
tools coming from broadly two different sets of techniques; this taxonomy
is borrowed from \cite{bunel_unified_2017}.

The first set is the family of exact verification methods, such as
Satisfiability Modulo Theory \cite{barrett2018satisfiability}. SMT solvers
perform automated reasoning on 
logical formulae, following a certain set of rules (a logic) on
specific entities (integers, reals, arrays, etc.) described by a theory.
An SMT problem consists in deciding 
%if,
whether,
for a given formula, there exists an instantiation of the variables 
that makes the formula true. 
Programs properties and control flow are encoded as logical formulae, that 
specialized SMT solvers try to solve. It is possible to express precise
properties, but since most SMT solvers try to be exhaustive over the 
search space, a careful formulation of the constraints and control flow is 
necessary to keep the problem tractable.
It was formally proven in 
\cite{katz_reluplex:_2017} that solving 
a verification problem composed of conjunction 
of clauses by explicit 
enumeration for a feedforward network is 
NP-hard. However, the NP-hardness of a problem does not prevent us from designing
solving schemes. In this same work, the authors introduced ReLuPlex, 
a modified solver and simplex 
algorithm, that lazily evaluates ReLUs, reducing the need 
to branch on non-linearities. Their follow-up work \cite{dillig_marabou_2019}
improves and extends the tool to support more complex networks and 
network-level reasoning.
Others \cite{bunel_unified_2017} rephrase the problem of adversarial  
robustness 
verification as a branch-and-bound problem and provide a solid benchmark
to compare current and future algorithms on piece-wise linear networks.
Other exact techniques are based
on mixed  integer linear programming (MILP).
%\cite{tjeng_evaluating_2017}. 
%The authors propose a formulation of
The verification of adversarial robustness properties on piece-wise linear 
networks
can indeed be formulated 
as a MILP problem \cite{tjeng_evaluating_2017}, and a 
pre-conditioning technique drastically reduces the number of necessary 
calculations. %They were able to check a
Adversarial robustness properties were thus checked
on ResNets (a very deep architecture) 
with $l_{\infty}$-bounded perturbations on CIFAR-10.

The second set of techniques in formal methods is based on
overapproximating the program's behaviour.
Indeed, since solving the exact verification problem is hard, some authors 
worked on computing a lower bound of $\varepsilon$, %\delta_s$,
using techniques 
building overapproximations of the program, on which it is easier to 
verify properties.
Abstract interpretation (first introduced in \cite{cousot_abstract_1977}) is 
an example of such technique.
It is a mathematical framework aiming to prove sound properties on 
\textit{abstracted semantics} of program. 
In this framework, a program's concrete executions 
are abstracted onto less precise, but more
computationally tractable abstract executions, using numerical domains.
Finding numerical domains that balance expressiveness,
accuracy and calculation footprint is one of the key challenges of abstract
interpretation.

The first instance of specifically-tailored 
deep learning verification described how to 
refine non-linear sigmoid activation function 
to help verification 
\cite{pulina_abstraction-refinement_2010}. 
\cite{wong_provable_2017} proposes an outer convex envelop for ReLu classifiers
with linear constraints, expressing the robustness problem as a Linear
Programming (LP) problem.
\cite{mirman_differentiable_nodate} and \cite{singh_abstract_nodate} propose a framework for building abstract interpretations of neural networks, which they use to derive a tight upper bound on robustness for various architectures and for regularization.
%Using their method, they proved a neural network to be robust at 94.18\% to adversarial samples bounded by a $l_{\infty}=0.1$ perturbation.
%\cite{mirman_differentiable_nodate} and \cite{singh_abstract_nodate} propose a framework for building abstract interpretations of neural networks, which they use to propose a tight upper bound on robustness precision for various architectures and for regularization.
On MNIST, both works displayed a robustness of 97\% bounded by a $l_{\infty}=0.1$ perturbation. 
On CIFAR-10,
they achieved a 50\% robustness for a similar net with a $l_{\infty}=0.006$ 
perturbation.
%\cite{wang_formal_2018} proposed to perform
Symbolic calculus on neural networks is performed in \cite{wang_formal_2018}, allowing symbolic analysis and outperforming previous methods.
A verification framework
based on bounding ReLu networks with linear functions is proposed in \cite{boopathy_cnn-cert:_2018}.

The boundary between these two families of techniques can be blurry, 
and both techniques can be combined. 
For instance, \cite{singh_robustness_2018} combines overapproximation 
and MILP techniques to provide tighter bounds on 
exact methods.  Competing with complete methods, they verify a hard property 
on ACAS-Xu faster and provide precise
bounds faster than other methods. 

All these techniques are employed either for proving local properties
(local adversarial robustness), or on simpler, non perceptual input spaces.
On the opposite, our work proposes a framework to prove global properties 
on perceptual inputs.

% Contrib 1: Formalism for specs
\section{\framework: a new formalism to specify and verify machine
learning models}
\subsection{Motivation}
In most deep learning application domains, such as image
classification \cite{he_deep_2015}, object detection \cite{chabot2017deep},
control learning \cite{bojarski2016end}, 
speech recognition \cite{ravanelli_pytorch-kaldi_2018},
or style transfer \cite{karras2019style}, 
there exists no formal definition of the input. Let us
consider the software of an autonomous car
as an example. A desirable property would be not to run over
pedestrians. This property can be split in i) all pedestrians are
detected, and ii) all detected pedestrians are avoided.
For a formal certification, the property should be expressed in the form ``For any image containing pedestrians, whatever the weather conditions or camera angle could be, all pedestrians present in that image are detected and avoided''.
Such a formulation supposes one is able to describe the set of all possible images containing pedestrians (together with their location).
However, there exists no
%universal  % no universal model is an issue also, but here the point is more about the existence even of such a model
exact characterization of what a pedestrian is or looks like,
and certainly not one that takes into account weather condition, camera angle,
input type or light conditions.
%The kind of property we aim for here could be phrased as ``For all images, all weather conditions, all pedestrians are detected and avoided''.
Any handmade characterization or model would be very tiresome to build, and
still incomplete. 

On the upside, machine learning has demonstrated its ability to make use of 
data that cannot be formally specified, yielding impressive
results in all above-mentioned application domains, among others; on the downside, 
it has also been demonstrated that ML models can easily
fail dramatically, for instance when attacked with adversarial examples.
Thus, manufacturers of critical systems need to provide elements that allow 
regulators, contractors and end-users to 
trust the systems in which they embed their software.

Usually, car manufacturers rely on test procedures to measure their system's
performances and safety properties.
But testing can, at best, yield statistical bounds on the absence of failures:
The efficiency of a system against a particular situation is not assessed 
before
this situation is actually met during a real-world experiment.
% Figure \ref{fig:base_case} shows a schematic representation of the pipeline we are concerned with here. 
As the space of possible situations is enormous (possibly infinite) and
incidents are rare events, 
one cannot assess that an autonomous vehicle will be safe in every
situation by relying on physical tests alone. 

% \begin{figure}[h!]
%     \centering
%     \includegraphics[width=0.8\columnwidth]{base_case.png}
%     % \includesvg[width=0.8\columnwidth]{base_case.svg}
%     \caption{Base case: a classifier learns on a training set
%         made of real world
%         data. No formal characterization for the inputs is available; 
%         only properties of the network or the output could be defined 
%     and checked.}
%     \label{fig:base_case}
% \end{figure}

A current remedy is to use artificial data, and to augment the actual data with data generated by a 
simulation software, with several benefits: Removing the need
to collect data with expensive and time consuming tests in the real world; 
Making it possible to generate potentially hazardous 
scenarios precisely, e.g., starting with the most common crash cases.
Examples of such simulators are Carla \cite{Dosovitskiy17} and the 
NVIDIA Drive Constellation system.
However, even if it is possible
to artificially generate corner cases more easily, the space of possible scenarios is still
enormous, and some accidents remain completely unpredictable \textit{a priori} by
human test designers.
For instance, in a recent car accident involving partially self-driving 
technology, the manufacturer admitted that the camera failed to distinguish a 
white truck against a bright sky \cite{hawkins_2019}, causing the death of the
driver. Such a test case is difficult to come up with for a human,
because it is the conjunction of specific environmental conditions and 
specific driving conditions. 

Our motivation is to bring an additional layer of trust, not relying on 
statistical arguments, but rather on formal guarantees. 
Our long term objective is to be able to formalize a specification and to
provide guarantees on every 
possible scenario, automatically finding violations of the specification.
Because practitioners are now relying more and more on simulators, we propose 
as a first step
to study such simulated setting, and to formalize it.
The idea is to rephrase the verification problem in order to include 
both the deep learning model \emph{and} the simulator software
within the verification problem.
As said earlier, a simulator offers more control on the learning data 
by providing
explicit parameters (for instance: number and positions of pedestrians on the
image). % \gyo{on parle du ML en général, on prend l'exemple de la voiture autonome. 
\subsection{Problem formulation and notations}
Let $f : $ \persp{} $\rightarrow$ \decsp{} be an algorithm taking a 
perceptual input $x \in$ \persp{} and yielding a decision $y \in$ \decsp{}. 
%\persp{} is the input space and \decsp{} the decision space.
%Typically, \persp{} will be the perceptual space,
The perceptual space \persp{} will typically be
of the form 
$\mathbb{R}^d$ or $[0,1]^d$.
In the general framework of this work, 
$f$ is a program
%learned
trained
with a learning procedure 
on a
finite
subset of \persp{} % \times \mathcal{Y}$} %% specific to supervised ML
to perform a specific task (e.g., drive the passengers safely home).
In our example, the task would be to output a command from an image, in 
which case, for a given image $x$, $f(x)$ would be the driving action taken 
when in environment $x$.

Let us denote by
\gen{} $:$ \prm{} $\rightarrow$ \persp{} the simulator, that is, a 
function taking as input a configuration $s \in$ \prm{} of parameters, and returning the result of the 
%associated
simulation
%, e.g., an image or video 
%with  pedestrians, weather, road, \dots, located and appearing
%as described by the
associated to these
parameter values.
%Let $s$ denote a configuration of parameters used as inputs to the simulator, 
%belonging to parameter space \prm, that
%can be  discrete or continuous variables.
A
%parameter
configuration
$s$ of parameters contains all information needed by the simulator to generate 
a perceptual input; each parameter may be a discrete or continuous variable.
Let us take as running example a simulator of autonomous car images: $s$ would
contain the road characteristics, the number of pedestrians and their positions, 
the weather conditions\dots, that is, potentially, thousands or 
millions 
of variables, depending on the simulator realism.

The problem to solve here is the following:
\textit{For a model $f$ trained on data belonging to \persp{} generated 
by \gen{} to perform a certain task, how can we formulate and formally 
verify practical safety properties for all possible $x \in $ \persp, 
including samples never seen during training}?

\subsection{Including the simulator in the verification}

\begin{figure}[h!]
    \centering
    \includegraphics[width=0.8\columnwidth]{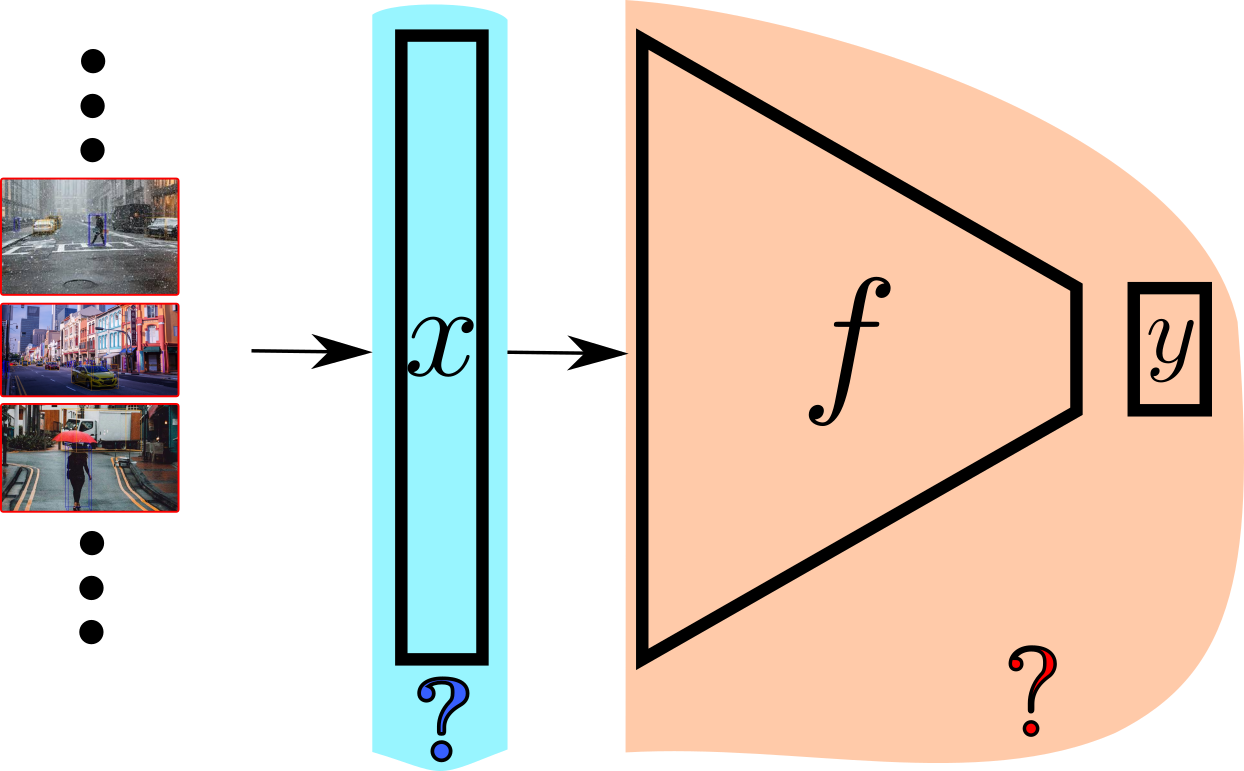}
    % \includesvg[width=0.8\columnwidth]{generated_case.svg}
    \caption{Natural inputs with huge perceptual space:
    no characterization of the input nor property can be formulated.}
    \label{fig:base_case}
\end{figure}

In standard settings,
such as the ones schematized in Figure~\ref{fig:base_case}, 
specifications express
relationships from \persp{} to \decsp{} using a formulation of $f$.
But \persp{} is such a huge space that formulating 
properties that are non trivial, let alone verify these, is prohibitively difficult, 
especially in the case of perceptive systems where the domain of $x$ cannot 
be specified: all matrices in $\left( [0,255]^3 \right)^{\#\mathrm{pixels}}$ 
are images, technically, but few of them make sense, and one cannot describe 
which ones. Moreover, given an image $x$, the property to check might be 
difficult to express, as, to state that all pedestrians were detected and 
avoided, one needs to know whether there are pedestrians in $x$ and where, 
which we do not know formally from just the image $x$. And if one had a way 
to retrieve such information from $x$ (number and location of pedestrians) 
without any mistake, one would have already solved the initial problem, 
i.e., safe self-driving car.

To summarize, in this setting, it is impossible to express a relevant space 
for $x$ and a property to verify $\Phi$:
  $$\forall x \in \,?, \;\; \Phi?\,\big( \, f(x) \, \big) \; $$

In the setting of simulated inputs,
% (described in figure \ref{fig:generated_case}), 
though it remains difficult to formulate properties on the perceptual space
\persp{}, we know that this space is produced by
\gen{} applied to parameters in \prm{}. On 
the contrary to 
\persp{}, \prm{} is a space 
where there exists an abstract, albeit simplistic characterization of
entities. Indeed, setting parameters for a pedestrian in the simulated
input yields a specification of what a pedestrian is in \persp{} according
to the inner workings of \gen{}. The procedure \gen{} transforms elements
$s \in$ \prm{}, that represent abstracted entities, 
into elements $x \in$ \persp{}
that describe these entities in the rich perceptual space.
To output values in \decsp{}, $f$ has to capture the inner semantics 
contained in 
\persp{}, that is to say, to abstract back a part of \prm{} from \persp{}. 

% \begin{figure}[h!]
%     \centering
%     \includegraphics[width=0.8\columnwidth]{generated_case.png}
%     % \includesvg[width=0.8\columnwidth]{generated_case.svg}
%     \caption{Generated inputs: a program generates some part of the 
%         training set.Without considering the simulator, 
%         there is no formal characterization of the input.}
%     \label{fig:generated_case}
% \end{figure}

The above remark is the key to the proposed framework: If we include \prm{} and \gen{} alongside $f$, \persp{} and \decsp{} in the
verification problem, then all meaningful elements of \prm{} are de facto included. 
It then becomes possible to formulate interesting
properties, such as ``given a simulator that defines pedestrians as
a certain
pattern
%configuration
of pixels, does a model
%learned
trained
on the images
generated by this simulator avoid all pedestrians correctly?''.
Formally, to ensure that the output $ y = p \circ{} g (s)$ satisfies a
property $\Phi$ for all examples $x = g(s)$ that can ever be generated by 
the simulator, the formula to check is of the form:
\[
\forall s \in \mathcal{S}, \;\; \Phi\big(s, \; p \circ{} g (s) \, \big) \;
\]
The property $\Phi$ may depend on $s$ indeed, as, in our running example, $s$ explicitly contains the information about the number of pedestrians to be avoided as well as their locations.

\begin{figure}[h!]
    \centering
    \includegraphics[width=0.7\columnwidth]{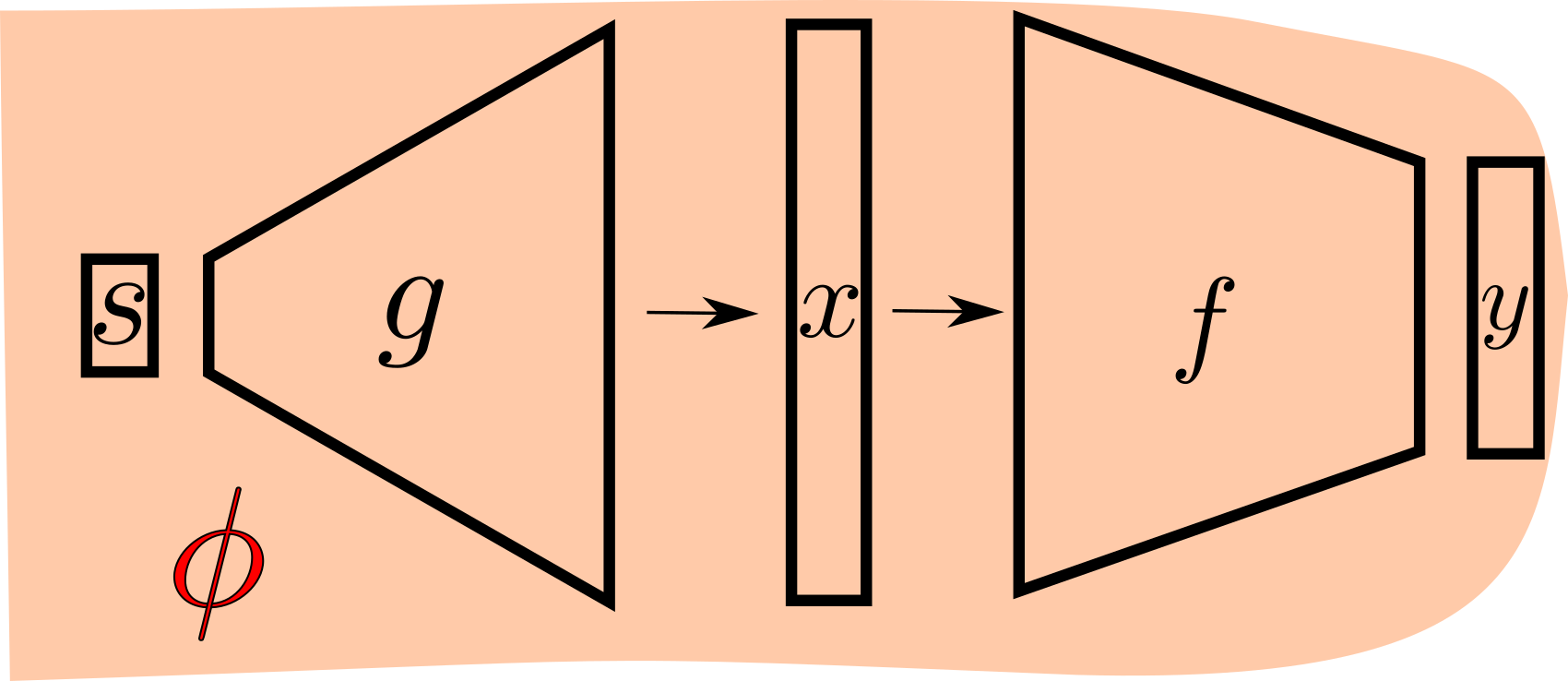}
    % \includesvg[width=0.8\columnwidth]{./generator_in_spec.svg}
    \caption{Generated inputs with integration of the generation procedure in
        the verification problem. There are now new properties to check 
        since we have a formal characterization of the perceptual elements.}
    \label{fig:generated_in_spec_case}
\end{figure}

Including \prm{} and \gen{}
in a formal property to check
requires to formulate at least partially 
the multiple functions that compose \gen{}. Describing precisely these
procedures is a key problem that we plan to address later. % that we will try to address later?

As our framework relies on including the simulator in the verification problem,
we call it Certifying Autonomous deep Models Using Simulators (\framework).
% For instance, if we have 
% two parameters: $s_1$, describing if there is a pedestrian, and $s_2$,
% describing where it is, one must know how a pedestrian is represented
% by \gen{} and how it is displayed on the image. 
% \gyo{[pas sûr d'avoir compris ce qu'apporte cet exemple / cette phrase. Plutôt:], a possible property to check for a pedestrian detector task would be that the number $y$ of pedestrians found matches the number $s_1$ indicated in the scenario given to the simulator [ou peut-être après la phrase suivante?]}. 
% A visual description of our formulation is shown figure
% \ref{fig:generated_in_spec_case}.

% \gyo{[NB: expliquer le code couleur des figures qqpart]}

\subsection{Separating perception and reasoning}

Before the rise of deep learning, the perception function 
(which, \eg recognizes
a certain
%configuration
pattern
of pixels as a 
pedestrian) and the control, or reasoning function 
(which, \eg{} analyzes the location of a pedestrian and 
proposes a decision accordingly) 
in vehicles were designed and optimized separately. However, work such as  
\cite{bojarski2016end} showed that end-to-end learning 
can in general be a much more efficient alternative; there exist
many incentives to adopt
this end-to-end architecture, mixing and training jointly the perception 
and control functions. However, combining
%the two
perception and reasoning 
into one model makes the formulation of safety properties more difficult.

\begin{figure}[h!]
    \centering
    \includegraphics[width=0.8\columnwidth]{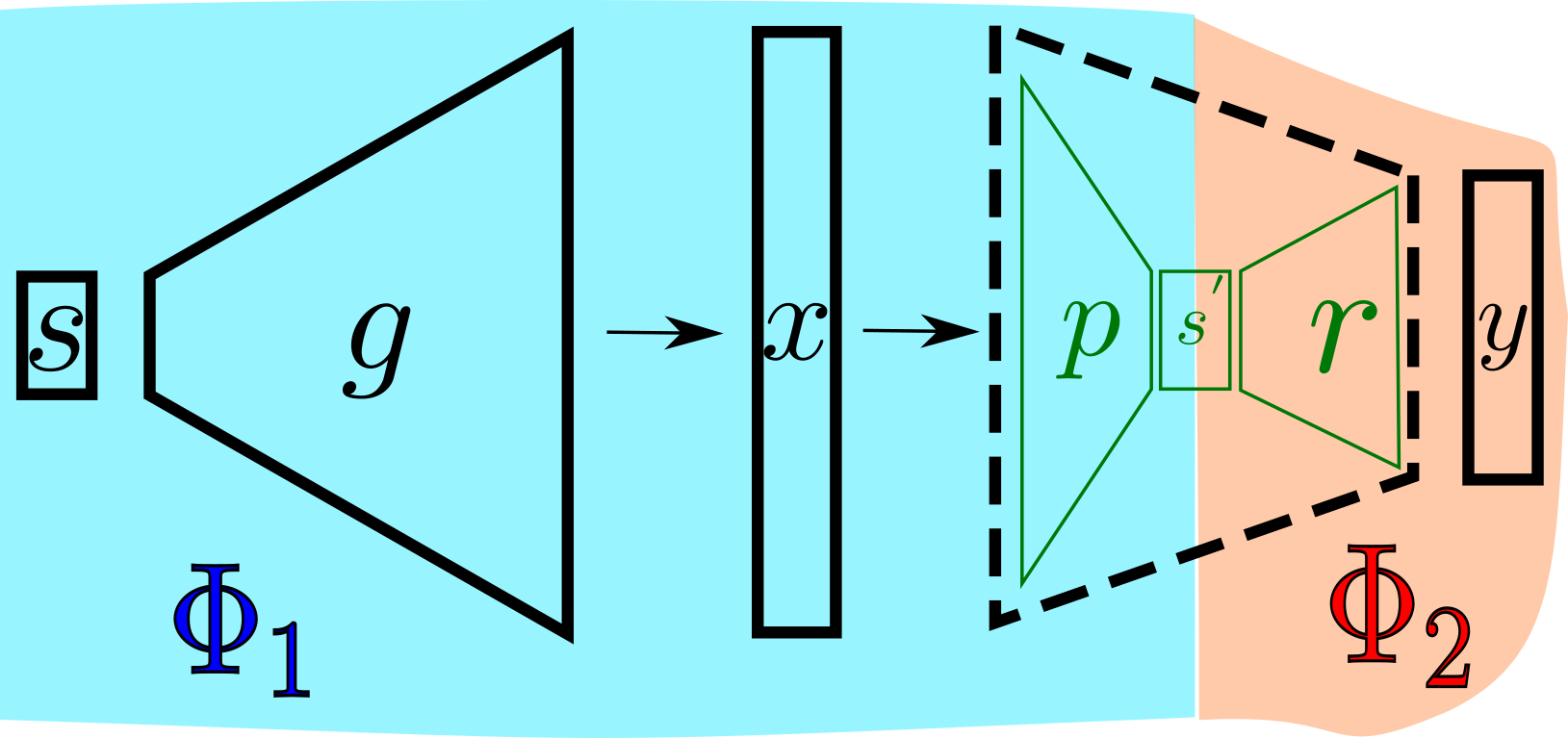}
    % \includesvg[width=0.8\columnwidth]{./generator_in_spec_sep.svg}
    \caption{Integration of the generation procedure in
        the verification, with split between perception and reasoning:
        \per{} learns to capture all the 
        relevant parameters; \rsn{} learns to respect the 
        specification. Verifying $\phi_1$ proves the perception
        unit once and for all; verifying $\phi_2$ can be done 
        when the specification changes (e.g., for different driving rules).}
    \label{fig:generated_in_spec_sep_case}
\end{figure}

Thus in our description (see Fig. \ref{fig:generated_in_spec_sep_case}), we choose to separate the perception and the 
reasoning functions, respectively in the components \per{} and \rsn{}. 
The perception part \per{}
is
%a classifier   %% bah non, plutôt régression en fait pour les paramètres continus
in charge of capturing all relevant information 
contained in the image, while the reasoning part \rsn{} will make use of this relevant
information to output directives accordingly to a specification.

One way to make sure that \per{} retrieves \emph{all} relevant information 
is to require it to retrieve \emph{all} information available, that is, to 
reconstruct the full simulator parameter configuration $s$. In this setting, 
the output $s'$ of the perception module \per{} lies in the same space as the 
parameter configuration space \prm, and the property we would like to be 
satisfy can be written as $ p \circ{} g = \mathrm{Id}$, 
which can be rewritten as:
\begin{equation}
  \label{eq:identity}
  \forall s \in \mathcal{S}, \;\; p \circ{} g (s)=s
\end{equation}
This way, we ensure that the perception module  \per{} correctly perceives \emph{all} samples that could ever be generated by the simulator.
In the case some parameters are known not to be relevant (image noise, decoration details, etc.), one can choose not to require to find them back, therefore asking to retrieve only the other ones. For the sake of notation simplicity, we will here consider the case where we ask to reconstruct all parameters.

%The learned output space of \per{} is a set \lprm{}, which is a     %% Attention aux confusions s' <> S'
%reconstruction of \prm{}, 
%that is to say for all $s \in $ \prm{}, $p \circ{} g (s)=s$. 
%Values $s' \in$ \lprm{} are representing a parameter configuration just as
%values $s$ do $\in$ \prm{}. 
%Phrased otherwise, the ideal goal for \per{} is to 
%output a space \lprm{} such that~ \lprm~=~\prm.
%\rsn{} learns the decision space \decsp{} using values from \lprm{} as inputs.
%% \gyo{cette phrase n'est vraiment pas claire, il faut vraiment développer. ``a set''?? part of a set? s'agit-il de retrouver tous les paramètres formant une configuration s, un sous-ensemble de ceux-ci, ou juste exprimer cette information mais pas forcément sous la forme d'un vecteur de paramètres identiques? Est-ce que s et s' vivent dans le meme espace? dans ce cas S = S'? (en tant qu'espaces) Je discuterais la possibilité d'extension / généralisation au cas ou S' est ``autre chose'' que S ou un sous-espace de S dans les remarques de la subsection suivante plutôt que ici si c'est ça ton but (rester général pour inclure d'autres propriétés que ``Id''). D'ailleurs je n'ai vu ``Id'' écrit nulle part, pas de formule mathématique ``quelques soit... , p o g(s) = s'' mise bien en évidence, ça manque vraiment pour y voir plus clair!}
% remarque prise en compte et modifications apportées.

%This separation allows us to formulate an additional property, more precisely,
%ensuring that \per{} has correctly captured all the relevant information 
%contained in \persp{}.
This separation between perception $p$ and further reasoning $r$ brings modularity as an additional benefit:
%An additional benefit is the modularity it brings:
even when dealing with different traffic regulations, it is only necessary to prove 
\per{} once; the verification of compliance towards
local legislations and specifications by \rsn{} can be done separately.
It allows to reuse the complex perception unit
with different reasoning modules $r$ without needing to re-prove it.
%, once it has been proven correct once. 
Note also that \rsn{} does not need to be as complex as \per{},
since it will work on 
much smaller spaces; multiple verifications of \rsn{} are then easier.
% See Figure~\ref{fig:generated_in_spec_sep_case} for a  visual representation of the complete framework.

One could argue that this formulation makes the problem more complex, and it
indeed may be the case. However, our proposition is aimed at safety, 
and in order to provide additional trust, 
it is sometimes necessary to formulate the problem differently.
For instance, there are good practices to structure the code to provide
some safety guarantees: bounded loops, correctly allocated and de-allocated
references, ban of function references, \dots~are constructs that voluntarily 
restrain the expressive power of the programming language to ensure a safer
behaviour.
Hence, although this formulation might 
seem like a step back w.r.t. the state-of-the-art, we argue that it provides 
a new way to formulate safety properties, and hence will be beneficial in the long run.

\subsection{Properties Formulation}
Considering jointly the simulator \gen{} and the ML model $f$, split in  \per{} and \rsn{}, two families of properties are amenable to formal checking:
%\smallskip
\begin{itemize}
%\noindent
\item $\Phi_1$: perception unit \per{} has captured sufficient
        knowledge from \persp{};
\item $\Phi_2$: reasoning unit \rsn{} respects a specification property
        regarding \decsp{}.
\end{itemize}
%\smallskip
% Regarding $\Phi_2$, the literature on deep learning verification displays some
% encouraging results on how to formulate and verify a verification problem: 
% lazy evaluation of ReLUs and translation as an SMT 
% problem \cite{katz_reluplex:_2017},\cite{dillig_marabou_2019} 
% formulation as a MILP \cite{tjeng_evaluating_2017}, or LP
% \cite{weng_towards_2018}, \cite{boopathy_cnn-cert:_2018} problem.
% Since deep neural networks use simple programming concepts (no
% loops), it is quite easy to translate them directly to a standard verification
% format, such as \smtlib{} \cite{barrett_smt-lib_nodate}. 
% Providing the inputs are 
% sufficiently well defined, it is then possible to encode safety
% properties as relationships between inputs and outputs, such as inequality constraints
% on real values, using the \lstinline{QF_NRA} theory of
% % for now we work on real values, we will switch when our tool will be able
% % to write on non scientific floating point values
% %on floating-point values, using of the \lstinline{QF_NRA} theory of
% \smtlib. 
Type 2 properties have been addressed in the literature - 
see Section \ref{reluplex}.
The key point of the proposed approach is thus to obtain
a representation space that 
reliably yields semantic meaning, which is the objective of 
$\Phi_1$. 
Since the simulator is included in the verification problem, 
properties of family $\Phi_1$ can be written as relationships between 
%elements of \prm{} and elements of \lprm.
input parameter configurations $s \in$ \prm{} and retrieved parameter configurations $s' \in$ \prm{}, 
outputs of the perception module $p$.
Strict equality between $s$ and $s'$ may be difficult to achieve, and is
actually not needed as long as the reasoning module $r$ is able to deal with
small estimation errors. 

%A soft version of property \ref{eq:identity} can be formulated as: 
%\emph{for all $s \in$ \prm{}, denoting $s' = p \circ{} g (s)$, 
%is $s'$ close enough to $s$?}

Expressed in the proposed
formalism, the perception task is equivalent to finding (a good approximation
of) \prm. Thus, a relaxed version of property \ref{eq:identity} to satisfy could 
be formalized as some tolerance $\varepsilon > 0$ on the reconstruction error $\|s'-s\|$ (for some metric $\|\cdot\|$):
\begin{equation}
  \label{eq:approx}
  \forall s \in \mathcal{S}, \;\; \left\| s - p \circ{} g (s) \right\| \leqslant \varepsilon
\end{equation}

\subsection{Discussion}
%As said before, learning all the parameter space is not always necessary.
As stated earlier, it is not always necessary to retrieve all parameters of configuration $s$.
For instance, one could seek to retrieve only the correct number of pedestrians and their locations, from any image generable by the simulator.
In this case, the output of \per{} would be
%different from
just a few coefficients of
$s$, and must
consequently
be characterized differently (\eg{} as belonging to a given subspace of \prm{}). This would 
allow to express more
%complex
flexible
properties than simply reconstruct all 
parameters.

For the model $f$ to correctly generalize, the simulated data must yield
two characteristics:
\begin{enumerate}
    \item they need to be sufficiently realistic 
        (that is to say, they should look like real-world images); if not the network
        could overfit the simplistic representation provided by the 
simulator;
        %        generator
    \item they must be representative of the various cases the model has to 
        take into account, to cover sufficiently diverse situations.
\end{enumerate}

Additional characterization of the simulator would be difficult. For instance, one could suggest to require the simulator $g$ to be either surjective or injective, in order to cover all possible cases $x \in$~\persp{}, or for parameters to be uniquely retrievable.
%with respect to a certain subspace of \persp{}.
Yet, the largest part of the perceptual space \persp{} is usually made of nonsensical cases (think of random images in $\left( [0,255]^3 \right)^{\#\mathrm{pixels}}$ with each pixel color picked independently: most are just noise), and the subspace of plausible perceptual inputs
%That subspace would represent the subspace
%of all images that can be generated by our generator and relevant to the 
%problem at hand. But in the general case, there is no characterization 
%of this hypothetical subspace; else the problem of safe self-driving would be 
%solved already.
is generally not characterizable (without which the problem at hand would already be solved).
Regarding injectivity, being one-to-one is actually not needed when dealing with properties
such as \ref{eq:approx}.

Finally, let us consider the case where several simulators are available, and
where, given a perceptive system $p$, we would like to assert properties of type $\Phi_1$ for each of the simulators.
% when we want to assert properties $\Phi_1$ on different simulators.
At first glance, as the output of $p$ consists of retrieved parameters, this would seem to require that all simulators are parameterized exactly identically (same \prm{}).
%A straightforward but tedious process would consist
%of encoding separately the generators and formally verify our model on both. 
%Note however that there is no need to retrain our model. 
%However, this process can be made easier if there exist a mapping between
%the parameters space of the two generators. It is even more easier if 
%one would like to only reconstruct a subspace of parameter (again, number and
%position of pedestrians). 
However, for real tasks, one does not need to retrieve all parameters but only the useful ones (\eg{} number of pedestrians and their locations), which necessarily appear in the configuration of all simulators.
As it is straightforward to build for each simulator a mapping from its full list of parameters towards the few ones of interest, a shared space for retrieved parameters can be defined, a unique perception system $p$ can be trained, and formal properties for all simulators can be expressed.

% Contrib 2: Tool for NN translation
% \section{ONNX2SMT: a tool to translate neural networks into logical formulae}
\section{Translating neural networks into logical formulae}

In previous section, we introduced two families of properties:
%, $\Phi_1$and $\Phi_2$:
$\Phi_1$ involves
the
%generator
simulator
$g$,
its parameter space \prm{},
%its output space, that is,
the perceptual
%generated input
space $\mathcal{X}$ where simulated data lie, and the perception unit \per;
$\Phi_2$ involves the representation space learned by \per (which should be a copy of \prm{}), the reasoning unit
\rsn{} and its output space $\mathcal{Y}$.

In order to be able to actually formulate properties of these families,
we must first be able to represent all these elements as logical formulae.
The goal of this section is to introduce \outil,
a tool to do so automatically.

\outil{} provides an
interface to all machine learning models that use the Open Neural
Network Exchange format \cite{bai2019}, and translates them into the
standard language \smtlib \cite{barrett_smt-lib_nodate}, allowing all 
state-of-the-art generalist SMT solvers and deep learning verification
specialized tools to work on a direct transcription of state-of-the-art neural
networks. \outil\ will be open-sourced to further help the community
effort towards safer deep learning software.

\subsection{ONNX and SMT-LIB}
The Open Neural Network eXchange format
(ONNX)\footnote{https://onnx.ai/} is a community initiative
kickstarted by Facebook and Microsoft, that aims to be an open format for
representing neural networks, compatible across multiple frameworks.
It represents neural networks as directed acyclic graphs, each node of
the graph being a call to an operation. Common operations
in machine learning and deep learning are tensor multiplications,
convolutions, activations functions, reshaping, etc.
\footnote{Full list of supported operators is available at:\\ \mbox{} \hfill
https://github.com/onnx/onnx/blob/master/docs/Operators.md}.
Operations have predecessors and successors, describing the flow of information
in the network.
The network parameters are also stored in the ONNX graph.
A wide variety of deep learning frameworks support ONNX, including Caffe2,
PyTorch, Microsoft CNTK, MatLab, SciKit-Learn and TensorFlow.
Examples of use cases presented in the official
page\footnote{https://github.com/onnx/tutorials/}
include benchmarking models coming from different frameworks, converting a
model prototyped using PyTorch to Caffe2 and deploying it on embedded software.

SMT-LIB2 is a standard language used to describe logical formulae to be
solved using SMT solvers. Most state-of-the-art solvers implement a
\smtlib{} support, which facilitates benchmarks and comparisons between 
solvers. SMT-COMP \cite{barrett2005smt} is a yearly competition using \smtlib{}
as its format. This challenge is a unique opportunity to present different
techniques
used by solvers, to increase the global knowledge of the SMT community.
SMT-LIB2 supports expressing formulae using bit vectors, Boolean operators,
functional arrays, integers, floating points and real numbers, as well as 
linear and non-linear arithmetic. In this work, only
the Quantifier-Free Non linear Real Arithmetic
(\lstinline{QF_NRA}) theory will be used.
Since the language aims to be compatible with a wide variety of solvers,
expressivity is limited compared to languages such as Python, used by most deep learning platforms.
%, powered by PyTorch.
In particular, there is no built-in Tensor type, and it is hence 
necessary to adapt the semantics of tensors to SMT-LIB2. This adaptation is 
%also supported
performed
by \outil.

\subsection{Features}
Features of \OUTIL\ include the support of the most common operations in modern
neural networks, such as tensors addition and multiplication, maxpooling and
convolution on 2D inputs. Support for a wider range of operators 
(such as reshaping or renormalization operators) is on-going work.

\OUTIL{} uses a Neural IntErmediate Representation (NIER) to perform 
modifications of the deep neural network structure, for instance by following
rewriting rules. NIER is still at an early stage, but future work will
integrate state of the art certified reasoning and pruning.

The conversion from ONNX to NIER is performed thanks to the
reference protobuf description of ONNX, converted to OCaml
types using the piqi\footnote{https://github.com/alavrik/piqi} tool suite.
\OUTIL\ %also
provides straightforward conversion from ONNX to SMT-LIB,
using NIER as an intermediate representation.

All the features described above
allow us to encode machine learning models
(\per and \rsn) as SMT formulae.
$\mathcal{X}$ and $\mathcal{Y}$ can be expressed directly
using \lstinline{QF_NRA} existing primitives.
Future work will provide an additional mechanism to encode
\gen{} and \prm{}.

\subsection{Usage}

% Figure \ref{fig:workflow} is a visual representation of
\OUTIL\ workflow can be summarized as follows:\\  
{\bf Input}: an ONNX file created using an ML framework;\\
1. Convert the ONNX model to NIER (\lstinline{onnx parser});\\
2. Convert NIER to a SMT-LIB (\lstinline{smtifyer}) string, written on disk;\\
3. Add the property to validate to the existing SMT-LIB file;\\
{\bf Output}: An SMT-LIB file that can be solved to prove the property.
    %  (integration of popular SMT solvers is planned).
% \end{enumerate}

% \begin{figure}[h!]
%     \centering
%     % \includesvg[height=0.7\textheight]{./imgs/workflow.svg}
%     \includegraPhics[height=0.7\textheight]{imgs/workflow.png}
%     \caption{Schematic workflow of our tool. Courtesy of
%     \cite{oord_wavenet:_2016}, \cite{he_deep_2015}, Wikipedia and
%     \cite{simonyan2014deep} for the pictures describing the models}
%     \label{fig:workflow}
% \end{figure}

% \todo{open source! maybe package an open source smt solver
% with the tool?}\zak{Il faut parler à florent de la licence}

% \zak{Vendre l'outil comme un inter-standard
% \\
% contextualiser, parler de la nécessité de traduire f et Phi et g et s
% \\
% parler d'ONNX, représentation intermédiaire
% \\
% diagramme workflow et interopérabilité: plein de flèches vers onnx, une flèche
% vers nier graph, plein de flèches vers solveurs smt
% \\
% fonctionnalités de base: traduction, broadcast, opérateurs gérés
% \\
% exemple pour parler du broadcast: petite matrice 2x2
% \\
% figure 4: ligne rouge pour signifier la zone de danger}

% \todo{Modify basic generator to take parameters a network can learn, for
% instance position of each pixel in the grid (vector of nxn)}

% Experimental settings, parameters and results
\section{Experiments}
% \todo{pour l'expérience, prendre un range pour noir et blanc, assigner à chaque
% pixel une probab d'être noir ou blanc}
As a proof of concept for the proposed framework, 
it is applied it to a simple synthetic problem.
%, and ACAS-Xu properties.
All experiments are conducted using the PyTorch framework
\cite{paszke2017automatic}. Neural networks are trained with PyTorch, then
converted into ONNX using the built-in ONNX converter and finally converted into
the intermediate representation in \smtlib{} format with \OUTIL.
We use z3\cite{deMoura2008}, CVC4\cite{barrett2011}, 
YICES\cite{Dutertre06theyices}
and COLIBRI\cite{marre2017real} SMT solvers as 
standard verification tools. 

Let us consider here the perception unit of an autonomous vehicle, 
whose goal is to output driving directives that result in safe 
driving behaviour. 
The perception unit is modeled as a deep neural network with one output node, taking as input an image.
If an obstacle lies in a pre-defined ``danger zone'', the network should 
output a ``change direction'' directive. 
Otherwise, it should output a ``no change'' directive. 

The ``simulator'' is here a Python script, taking as input the number and the
locations on the image of the one-pixel wide obstacles
and generating the corresponding black-and-white images. %on the screen. 
% Each neural network is tested against a validation set, generated with the same
% simulator used to generate the training data. There is no overlap between training
% and validation data.

The verification problem consists in the formulation of the network structure 
and constraints on the inputs, and in the following properties to check:
\begin{enumerate}
    \item verify that an input with an obstacle (or several ones) in the danger zone 
        will always lead to the ``change direction'' directive;
    \item verify that an input without obstacle on the danger zone 
        will never lead to the ``change direction'' directive.
\end{enumerate}
If both properties are verified, our model is perfect for all the inputs that 
can be generated. 
If the first one is not verified, %it provides
our verification system will provide
examples of inputs 
where our model fails, which can be a useful insight on the model flaws. Such
%inputs can also
examples could
be used for further more robust training,
%where they are
\ie{} integrated into a
future training phase to correct the network misclassifications. 
Similarly, if the second property is not verified, %it provides 
the solver will provide
false positives, that can help
designers reduce erroneous alerts
%, making
and make their tools more 
acceptable for the end-user. % as well as correcting their programs flaws.

\subsection*{Experimental setting}
\begin{figure}
    \centering
    % \includesvg[width=0.4\columnwidth]{inputs_simple}
    \includegraphics[width=0.4\columnwidth]{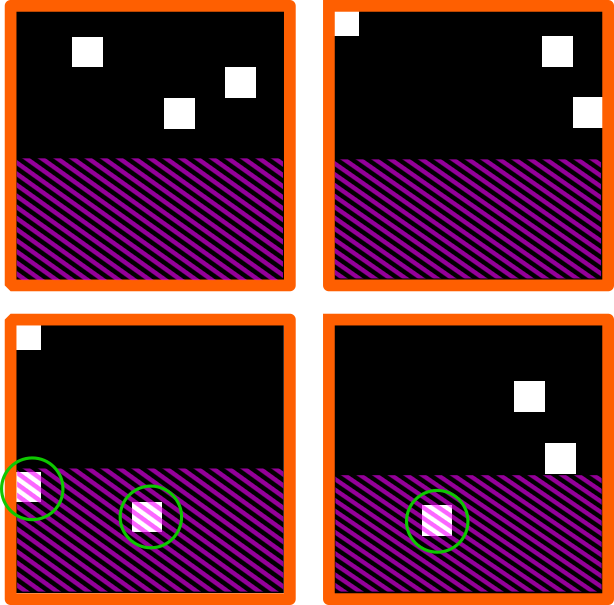}
    \caption{Example of inputs for the toy problem. White pixels represent
    obstacles. % The top of the image is far from the input sensor, while the bottom is near. The danger zone is dashed.
    If they are in the top half of the image, no alert should fire (first two exemples), while an alert must fire if
in the (dashed) bottom half of the image (last two examples). 9x9 picture is
depicted here for clarity.}
    \label{fig:simple_inputs}
\end{figure}

In this toy example, input data are $N \times N$ black-and-white images % whose pixels can only take black or white values. % Experiment is conducted with 9x9 input size.
(see Fig.~\ref{fig:simple_inputs} for examples). %an example of inputs.
The space of possible simulated data $g(\mathcal{S}) \subset \mathcal{X}$ 
can simply be described by the constraint that each pixel
%The resulting constraint on inputs is simply that each input
can only take 
two values (0 and 1). In real life,
%values are continuous and
data are much more complex, possibly continuous; such data can also be handled in our framework, though experimenting with %more
realistic simulators is the topic of future work.
%generalizing to such contexts is the topic of future work.
The neural network is fully-connected with two hidden layers.
The number of neurons in the first and second hidden layers
%has a number of neurons equal to half of
are respectively one half and one quarter of the flattened size of the input.
%the second hidden layer has a number of neuron equal to a quarter of the flattened size of the input.
All weights were initialized using Glorot optimization, with a
gain of 1. 
The network was trained with Adam optimizer for 2000 epochs, 
with batch size of 100, using the binary cross entropy loss.
The danger zone is defined as the bottom half part of the image.
Any image with at least one white pixel %on this half bottom
in this zone
should then % output
yield a ``change direction'' directive.

Here, constraints on inputs are encoded as statements on the \smtlib{}
variables. A fragment of property to check is presented on Figure
\ref{fig:simple_property}.
\lstset{basicstyle=\small\ttfamily,columns=fullflexible}
\begin{figure}[h!]
    \begin{lstlisting}[linewidth=\columnwidth,breaklines=true]
    ;;;; Automatically generated part
    ;; Inputs declaration
    (declare-fun |actual_input_0_0_0_8| () Real)
            [ommitted for brevity]
    ;; Weights declaration
    (declare-fun |l_1.weight_31_4| () Real)
    (assert (= |l_1.weight_31_4| (/ -5585077 33554432)))
            [ommitted for brevity]
    ;; An example of encoded calculation
    (assert (= |8_0_0_0_39| (* |actual_input_0_0_0_8| (+ |7_80_39| (* |actual_input_0_0_0_7| (+ |7_79_39| (* |actual_input_0_0_0_6| (+ |7_78_39| (* |actual_input_0_0_0_5|     
            [ommitted for brevity]
    ;; Outputs declaration
    (assert (= |actual_output_0_0_0_1| ( + |16_0_0_0_1| |l_3.bias_1| )))
            [ommitted for brevity]

    ;;;; Handmade annotations
    ;; Simulator description
    ;; Input space constraints: inputs between 0 and 1
    (assert (or (= actual_input_0_0_0_8 0) (= actual_input_0_0_0_8 1.)))
            [ommitted for brevity]
    ;; Property to check
    ;; At least one input in the danger zone is white
    (assert 
      (or
        (or (= actual_input_0_0_0_5 1.)
            (= actual_input_0_0_0_6 1.))
        (or (= actual_input_0_0_0_7 1.)
            (= actual_input_0_0_0_8 1.))
            [ommitted for brevity]
    ;; Formulate constraint on outputs:
    ;; Output is always higher than a 
    ;; confidence value
    ;; Negation: output can fire lower than a 
    ;; confidence value
    (assert (< actual_output_0_0_0_1 actual_output_0_0_0_0))
    \end{lstlisting}
    \caption{A SMTLIB2 file describing our problem.
    First part is a full description of the network,
    automatically produced by \OUTIL{}.
    Handmade annotations describe the property to check; the goal is to find 
a counterexample.}
    \label{fig:simple_property}
\end{figure}
On such a simple problem, the decomposition perception/reasoning 
is not needed,
%useful,
since there exists a formal characterization of what an obstacle is. 
%Thus we only seek to show that \OUTIL{} can effectively translate a model from ONNX to \smtlib.

Experiments results will come on later versions of the paper, and 
will additionaly be available at \href{https://www.lri.fr/~gcharpia/camus/}
{https://www.lri.fr/~gcharpia/camus/}.
% while it was trained on only a small sample of possible cases 
% (a few thousands, compared to $2^{81} \simeq 10^{24}$ possibilities).

%is never false for every possible input it was designed to take as input.
%Note that we learned on few examples compared to the number of possible 
%cases ($2^{81}$ for a 9x9 image with only binary valued pixels), 
%but we verified for all possible cases.

% Discussions and possible extensions
\section{Discussion and perspectives}
We introduced \framework, a formalism describing how to formally express safety properties 
on functions taking simulated data as %an
input. We also proposed \outil, 
a tool soon to be open sourced, that leverages two standards used by the 
communities of formal methods and machine learning, to automatically
write machine learning algorithms as logical formulae.
% note: pluriel de formula = formulas ou formulae mais pas formulaes
We demonstrated the
%use of \outil{} with the formalism
joint use of \outil{} and \framework{}
on a synthetic example mimicking a self-driving car perceptive unit, 
as a proof of concept of our framework.
This toy example is of course still simplistic and much work on scalability is needed before real self-driving car simulators can be incorporated 
into formal proofs.

%Several research tracks will be pursued.
Among future work, 
\outil{} will be released and gain support for more deep learning operations.
While we provide a toolkit to translate neural networks directly in our 
framework, a way to easily represent a simulator is yet to include.
It is not an easy task, since the simulator must be describable with 
sufficient granularity to allow the solver to use the simulator internal
working to simplify the verification problem. A scene description language 
with a modelling language for simulators
%would represent a credible
is a possible
answer to these issues. 
Further theoretical characterization of the simulator procedure and its link with the perceptive unit will be undertaken, for instance to encompass stochastic processes.
%Finally,
Besides,
on more complex simulators, programs and examples, the problem to
verify will remain computationally difficult. Various techniques to enhance 
solvers 
performances will be developed and integrated in \framework,
taking advantage of the domain knowledge provided by the 
simulators parameters.
Finally, our current framework
checks properties for all possible inputs,
including anomalous ones such as adversarial attacks.
%does not allow to distinguish genuine test errors from anomalous inputs
%that is to say, it relies on the assumption that the 
%classifier is perfect, at least on its training set.
A possible extension
would be to
%exhibit
identify
``safe'' subspaces instead, where
%the detection
perception
is guaranteed to be perfect,
and ``unsafe'' subspaces where
failures may happen.
%the detection is not guaranteed to be perfect.

\bibliography{biblio}

\end{document}